\documentclass[conference]{IEEEtran}
\IEEEoverridecommandlockouts
\usepackage{microtype}
\usepackage{graphicx}
\usepackage{subfigure}
\usepackage{booktabs}
\usepackage{amsmath}
\usepackage{amsfonts}
\usepackage{amssymb}
\usepackage{fontawesome}
\usepackage{hyperref}
\usepackage{url}

\def\BibTeX{{\rm B\kern-.05em{\sc i\kern-.025em b}\kern-.08em
    T\kern-.1667em\lower.7ex\hbox{E}\kern-.125emX}}
\begin{document}

\title{DeepFolio: Convolutional Neural Networks for Portfolios with Limit Order Book Data
\thanks{Sections I, II, III, IV were supported by the Ministry of Education  and  Science  of  the  Russian  Federation  (Grant no. 14.756.31.0001). Other sections were supported by the Mexican National Council for Science and Technology (CONACYT), 2018-000009-01EXTF-00154.}
}

\author{\IEEEauthorblockN{
Aiusha Sangadiev\IEEEauthorrefmark{9}\IEEEauthorrefmark{1},
Rodrigo Rivera-Castro\IEEEauthorrefmark{9}\IEEEauthorrefmark{2}, 
Kirill Stepanov\IEEEauthorrefmark{9}\IEEEauthorrefmark{3},\\ 
Andrey Poddubny\IEEEauthorrefmark{9}\IEEEauthorrefmark{4}, 
Kirill Bubenchikov\IEEEauthorrefmark{9}\IEEEauthorrefmark{5},
Nikita Bekezin\IEEEauthorrefmark{9}\IEEEauthorrefmark{6},\\
Polina Pilyugina\IEEEauthorrefmark{7} and
Evgeny Burnaev\IEEEauthorrefmark{8}
}
\IEEEauthorblockA{Skoltech\\
Moscow, Russia\\
\IEEEauthorrefmark{9}Equal Contribution
\\
Email: \IEEEauthorrefmark{1}aiusha.sangadiev@skoltech.ru,
\IEEEauthorrefmark{2}rodrigo.riveracastro@skoltech.ru,
\IEEEauthorrefmark{3}kirill.stepanov@skoltech.ru,\\
\IEEEauthorrefmark{4}andrey.poddubny@skoltech.ru,
\IEEEauthorrefmark{5}kirill.bubenchikov@skoltech.ru,
\IEEEauthorrefmark{6}nikita.bekezin@skoltech.ru,\\
\IEEEauthorrefmark{7}polina.pilyugina@skoltech.ru,
\IEEEauthorrefmark{8}e.burnaev@skoltech.ru
}
}
\maketitle

\begin{abstract}
This work proposes DeepFolio, a new model for deep portfolio management based on data from limit order books (LOB). 
DeepFolio solves problems found in the state-of-the-art for LOB data to predict price movements.
Our evaluation consists of two scenarios using a large dataset of millions of time series.
The improvements deliver superior results both in cases of abundant as well as scarce data.
The experiments show that DeepFolio outperforms the state-of-the-art on the benchmark FI-2010 LOB. 
Further, we use DeepFolio for optimal portfolio allocation of crypto-assets with rebalancing.
For this purpose, we use two loss-functions - Sharpe ratio loss and minimum volatility risk. 
We show that DeepFolio outperforms widely used portfolio allocation techniques in the literature.
\end{abstract}


\section{Introduction}
\label{submission}
\begin{figure}
%\vskip 0.2in
\begin{center}
\centerline{\includegraphics[width=0.7\columnwidth]{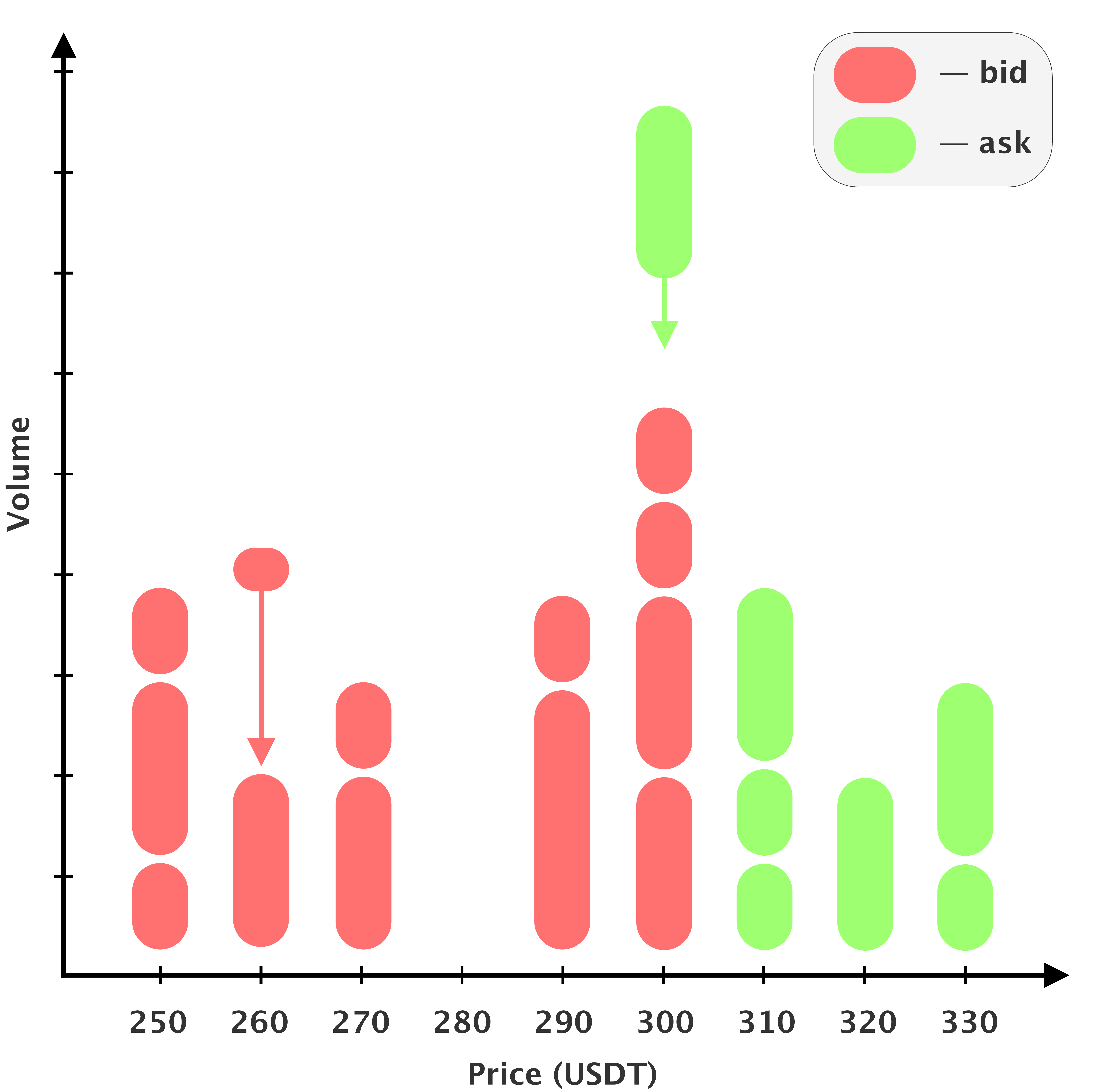}}
\caption{Example of a LOB for the ETH crypto-asset}
    \label{lob}
\end{center}
\end{figure}

More than half of the financial world uses electronic Limit Order Books (LOBs).
LOBS are a store of records of all transactions, \cite{rosu2010liquidity}, \cite{parlour2008limit}. 
A limit order is a request to transact with a financial instrument at a price not exceeding a threshold, \cite{murphy1986technical}. 
Usually, traders set so-called buy limit orders below the current market price.
They represent the maximum price that the trader is willing to pay. 
On the other side, traders set the amount above the current market price.
The sell limit orders act as the minimum price to sell. 
LOBs are also gaining popularity in the relatively new and rapidly developing crypto-asset market. 
The novelty of LOBs leads to low market liquidity and increased stochastic behavior of crypto-asset prices \cite{carrie2006new}.
It is easy to see the drivers behind the increasing popularity of LOBs.
Our example in \autoref{lob} shows how traders control the price of the transaction and the logic behind a LOB.
First, a passive order for one ETH crypto asset at 260 USDT arrives. 
Similarly, a retail order to sell three crypto-assets at 300 USDT appears.
The sell order matches with three passive orders to buy. 
Second, a trade executes at 300 USDT, and the LOB removes the buy orders.

Accordingly, modeling LOBs with mathematical methods is a challenging task. 
Typically, researchers resort to using the autoregressive integrated moving average model (ARIMA), \cite{ariyo2014stock}.
Alternatively, the vector autoregressive model (VAR), \cite{zivot2006vector}, is a popular choice.
One of the benefits of the VAR is that it can display the direction of transactions.
However, LOB data is highly stochastic, and time series are unsteady.
The result is additional noise to the data. 
This setting makes the creation of dedicated models and processing data demanding. 
Another limitation of those techniques is that they make assumptions on the data.
To overcome these limitations, \cite{zhang2019deeplob} suggests a state-of-the-art model called DeepLOB. 

In this work, we propose a LOB-based approach to predict price trends of crypto-assets.
Levering deep neural networks, we call our approach DeepFolio.
Our proposal achieves superior results and addresses some of the problems of DeepLOB. 
Moreover, we go a step further and use DeepFolio to build investment portfolios. 
Thus, this work adds a new entry to the "deep portfolio" literature. 

The remainder of the paper is structured as follows. 
We introduce relevant literature in \autoref{sec:lr} and our data processing in \autoref{sec:data}. 
We then proceed to present our methodology in \autoref{sec:model}. 
\autoref{sec:results} describes our experiments and details the results of our method. 
In this section, we compare with a range of baseline algorithms. 
In \autoref{sec:conclussion}, we summarise our findings and discuss possible future work.

\section{Related work} \label{sec:lr}
\subsection{Deep learning and LOBs}
There have been previous attempts to work with limit order book data using machine learning methods. 
For example, in \cite{passalis2018temporal}, they extract features using principal component analysis (PCA).
Furthermore, in a second step, they use linear discriminant analysis (LDA). 
However, these techniques are suitable only for processing statistical data. 
Besides, they are not optimized for working with dynamics. 
Another critical point is that these models make inherent assumptions about the data.
As a result, these techniques yield lower efficiency.

Besides \cite{zhang2019deeplob}, there are several works in the literature.
Their focus is on the application of deep learning and neural networks.
They use them to process limit order book data, and then to classify price trends. 
Along this line of work, one of the most notable entries is \cite{Tsantekidis2017}. 
The authors propose to use a fully convolutional neural network (FCNN).
With the FCNN, they extract features and perform trend classification. 
This approach shows significant improvements over more conventional methods such as support vector machines (SVM). 
Another example of using deep learning as a classifier for LOB data is \cite{Tsantekidis2017lstm}.
In this work, the authors applies an LSTM to perform trend forecasting based on LOB data. 
Finally, \cite{zhang2019deeplob} combines these two approaches to create a mixed CNN and LSTM neural network. 
With LOB data, the approach delivers state-of-the-art results in the classification of trends.
Significantly, it outperforms approaches using pure CNNs or LSTMs.

\subsection{Markowitz mean-variance model}
The Markowitz mean-variance model is a classic approach. 
Portfolio managers use it widely for portfolio building.
The central assumption underlying this theory is that the investor has two choices.
She will try to maximize profits at a given level of risk or minimize risk at a given level of profit. 
The Markowitz model offers to build a broad array of possible portfolios to reach these goals.
It then chooses one of them through the optimization of the risk-return curve.
To build the space of possible portfolios, Markowitz proposes to lever three elements.
It requires a class of assets, a vector of the average expected returns, and a covariance matrix, \cite{markowitz1978portfolio}. 
With this, the Markowitz model constructs an array of portfolios with various profitability-risk ratios, \cite{markowitz1978portfolio}.
Since the analysis builds on two criteria, the manager selects the portfolios based on three choices:

\begin{itemize}
    \item She searches for effective or non-improvable solutions.
    \item She chooses the main criterion, i.e., minimum profitability, using other criteria as constraints.
    \item She provides a "super criteria," such as a superposition of the previous two options.
\end{itemize}

In this work, the criteria for choosing the optimal portfolio are the maximum Sharpe Ratio, \cite{sharpe1966mutual}.
It is a standard metric for assessing the "optimality", and the minimum volatility risk.

\section{Data}\label{sec:data}
\subsection{FI-2010 dataset}
This dataset is the first public marked-up dataset of high-frequency financial markets, \cite{ntakaris2018benchmark}.
It is ideal for assessing and controlling the forecasting of indicators.
With time-series data from five stocks of the NASDAQ Nordic stock market, it consists of normalized representations.
It results in a dataset of approximately 40,000,000 time-series samples representing ten consecutive days. 
The dataset provides three different normalizations: z-score, min-max, and decimal precision normalization. 
Due to its richness and relevance, it is a good benchmark for LOB-based deep learning models, \cite{zhang2019deeplob}.

\subsection{Crypto-assets dataset}
Limit order books for crypto-assets are not readily available. 
Hence, we assemble the datasets using the public API of Binance, \cite{Binance}.
Binance is a relevant market for the trade of crypto-assets.
In our dataset, the time length of the collected data is one year.
It starts on February 27, 2019, and has an hour resolution. 
The data consists of orders, defined by bid or ask labels, time steps, volumes, and prices.
By asks and bids, we divide the orders.
We take the ten best asks, the ten best bids, and their respective volumes within a five-minute interval. 
As a result, we obtain 40 values for a single time step. 
Each of them consists of 20 asks and bids, as well as 20 volumes. 
The percentage of missing values is less than 6\%.
The dataset has missing values distributed evenly.
For data imputation, we consider methods relying on neighboring values. 
These are prices connected to an order volume, such as simple arithmetic or root mean square average.
However, it probably leads to a distortion of data. 
For this reason, we use the propagation of the last viable value as an additional imputation technique.
Moreover, we normalize the data using the dynamic $z$-normalization, see \autoref{Eq: z-score}. 

\begin{equation}
    z = \frac{x-\mu}{\sigma}
    \label{Eq: z-score}
\end{equation}

We use the mean $\mu$ and the standard deviation $\sigma$ of the previous five days.
The objective is to normalize the values of the current day. 
In the financial time series literature, dynamic normalization is a reasonable choice.
The motivation is that financial time series are usually affected by regime shifts, \cite{zhang2019deeplob}.
In particular, we can represent crypto-assets' prices as a sum.
For \cite{conrad}, the sum consists of the primary trend plus some noise or long term and short term volatility. 
Along these lines, the dynamic normalization enables the data to be within an appropriate range.
If we apply $z$-normalization on the whole dataset, we destroy the underlying data patterns.
Finally, for each point in the dataset, we establish a mid-price outlined in \autoref{Eq: mid-price}. 
It is the average between the best ask and the best bid. 
Throughout this work, we use mid-prices for further calculations.

\begin{equation}
    p_t = \frac{p^{(1)}_a (t) + p^{(1)}_b (t)}{2}
    \label{Eq: mid-price}
\end{equation}

After that, we generate three labels indicating price movements such as increase, decrease, or uncertainty. 
The third label is defined whenever an increase or decrease is too small to confirm them. 
Since financial data is inherently noisy and highly stochastic, we use label smoothing strategies. 
For this purpose, we calculate $m_-$, see \autoref{Eq: m_minus}, and $m_+$, see \autoref{Eq: m_plus}. 
These values denote the average of the previous and next $k$ mid-prices. 
We then calculate the "smoothed labels" $l_t$.
In \autoref{Eq: l_t pt} and \autoref{Eq: l_t m-} respectively, we outline these labels. 
These values show relative changes in the asset and its trend, taking into account a $k$-point smoothing.

\begin{equation}
    m_- (t) = \frac{1}{k} \sum^{k}_{i=0} p_{t-i} 
    \label{Eq: m_minus}
\end{equation}

\begin{equation}
    m_+ (t) = \frac{1}{k} \sum^{k}_{i=0} p_{t+i}
    \label{Eq: m_plus}
\end{equation}

\begin{equation}
    l_t = \frac{m_+ (t) - p_t}{p_t}
    \label{Eq: l_t pt}
\end{equation}

\begin{equation}
    l_t = \frac{m_+ (t) - m_- (t)}{m_- (t)}
    \label{Eq: l_t m-}
\end{equation}

For the final label distribution, we set a threshold, $\alpha$, equal to 0.001. 
Changes of 0.1\% are sufficiently large to indicate a price movement. 
If $l_t > \alpha$, we apply $l_t$ to signalize an increase.
Otherwise, if $l_t < -\alpha$, the price is decreasing.
We consider the $[-\alpha, \alpha]$ interval to be an intermediate value of $l_t$.
In this case, there is no increase or decrease in price.
The changes are insignificant for this range of values. 
We present this logic for the crypto-asset BTC.
In our example, the green background represents a buy signal.
We use red for the sell signal and white for the hold one.

\begin{figure}[!ht]
%\vskip 0.2in
\begin{center}
\centerline{\includegraphics[width=\columnwidth]{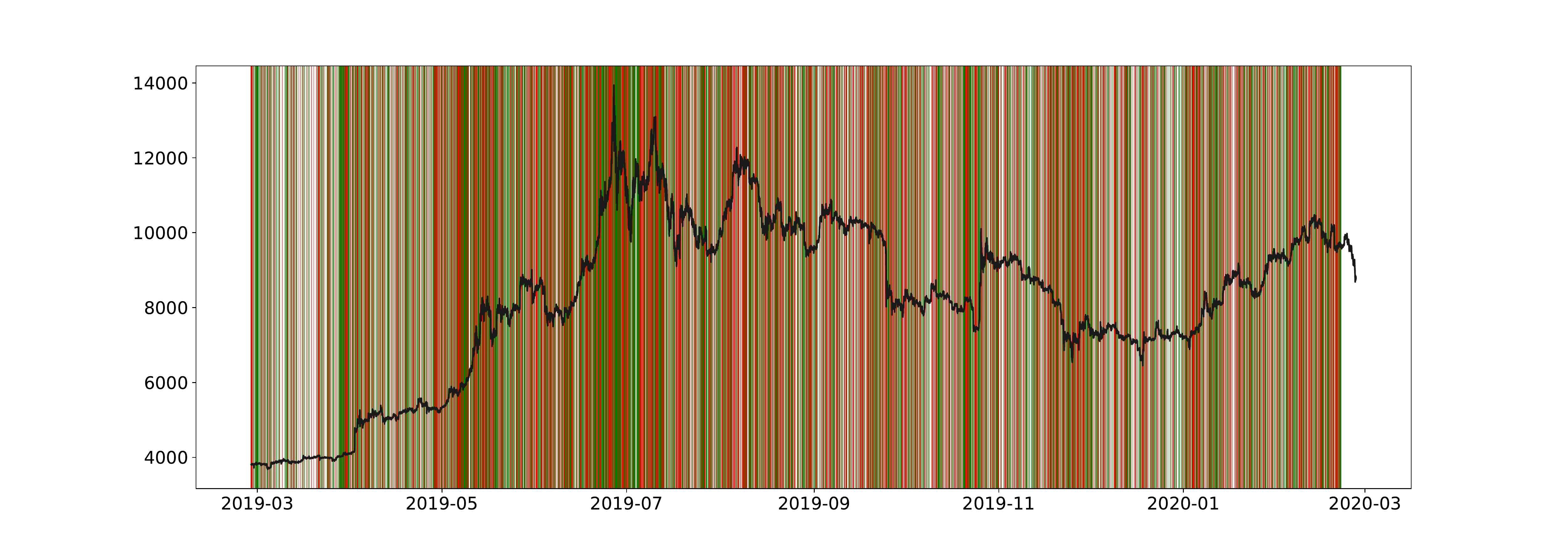}}
\caption{An example of the labeling process for the crypto-asset BTC. Green represents buy, red sell and white hold.}
    \label{btc_labels}
\end{center}
\end{figure}

\section{Model architecture}\label{sec:model}
\subsection{CNN+RNN}
The first module of DeepFolio consists of three main blocks.
The first block is a fully convolutional neural network (FCNN).
One Inception block represents the second block.
An LSTM network is our third one. 
The input to this network has three elements.
They are a batch size, a sequence length, and features. 
Hence, we consider this module to be a "CNN+RNN."

The FCNN block has three sub-blocks.
On the first block, we have a stridden convolutional layer.
It has a kernel size of $1 \times 2$.
Thus, it performs convolutions strictly over LOB levels. 
Two convolutional layers follow it in the second block.
Due to their kernel sizes of $4 \times 1$, they capture short-term time dependencies. 
In the last block of the FCNN, the kernel size expands to $1 \times 10$.
Hence, it performs convolutions over the remaining elements in the feature dimension.

Similarly, we employ an Inception block, \cite{szegedy2015going}. 
It enables us to capture dynamic behaviors over multiple time scales. 
This block is equivalent to performing multiple moving averages over different periods.
For financial time series analysis, we can use it to capture the time-series momentum.

The last LSTM block captures long-term temporal dependencies in the data. 
We feed its output into a fully-connected layer with a softmax activation function.
It has three outputs to produce probabilities of having one of three possible labels.
They are a negative price trend, $-1$, a neutral trend, $0$, and a positive trend, $+1$.

\subsection{Problems with the "CNN+RNN" module}
\paragraph{Extreme sensitivity to initial model weight allocation}
Empirical observations show that using "He uniform" is suboptimal.
Practitioners use it to initialize weights of the convolutional and recurrent layers, \cite{he2015delving}.
Nevertheless, both in the case of the weight matrices and the biases, the model "dies."
It happens early in the training process and results in a lack of learning. 
A better option is to use Glorot uniform, \cite{Glorot10understandingthe}.
It initializes the weight matrices of the CNN and the input weight matrix of the LSTM. 
Similarly, for the recurrent weight matrix of the LSTM, we have zero initialization.
We do this for all biases, and orthogonal, \cite{saxe2013exact}. 
In \autoref{exp1}, we show this effect. 
In the figure, "default" stands for the default initialization.
The second label, "initialization," represents our proposed allocation.

\paragraph{Slow learning process at the beginning of the training}
This effect is especially noticeable with the crypto-asset data.
Compared to the benchmark dataset, FI-2010, it is a smaller dataset. 
\autoref{exp2} depicts that it takes more than 30 epochs before proper training starts.

\paragraph{Worse depth-wise scalability}
It stems from the first two problems.
Unfortunately, the original model offers worse depth-wise scalability.
An increase in depth hampers the training process even further.

\subsection{ResCNN+GRU}
\begin{figure}[ht!]
%\vskip 0.2in
\begin{center}
\centerline{\includegraphics[width=\columnwidth]{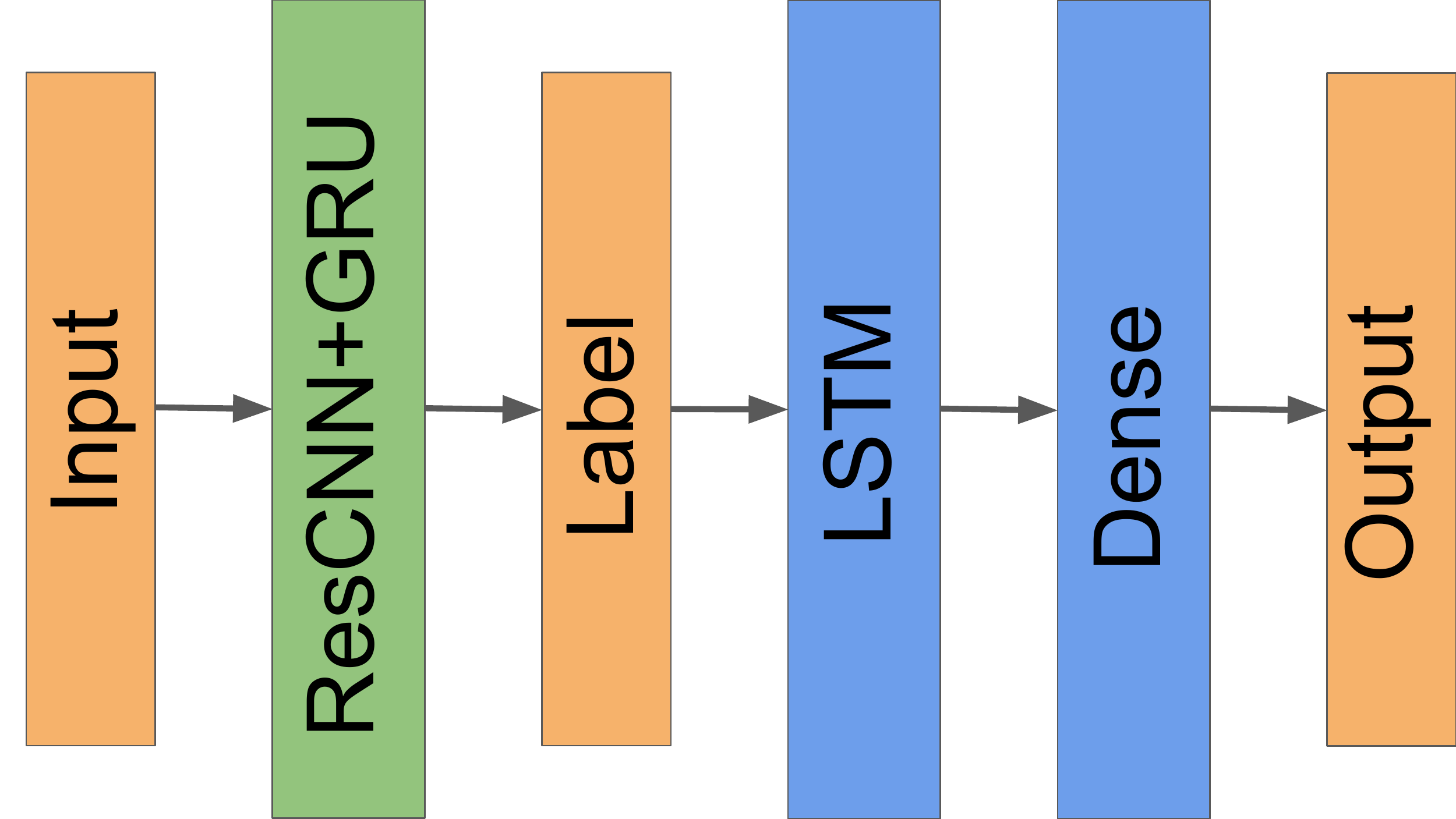}}
\caption{General architecture of DeepFolio}
    \label{deepfolio}
\end{center}
\end{figure}

In \cite{He2015}, the authors propose using residual connections.
The motivation is to improve the learning process of deep convolutional networks.
Residual connections allow for better gradient flows through the layers. 
Inspired by this, we introduce blocks with residual connections into the network. 
Our objective is to extend the depth of the network.
We also want to improve problems associated with gradient flows and vanishing gradients.
In \autoref{deepfolio}, we present a general architecture for DeepFolio.

\begin{figure}[ht!]
%\vskip 0.2in
\begin{center}
\centerline{\includegraphics[width=0.6\columnwidth]{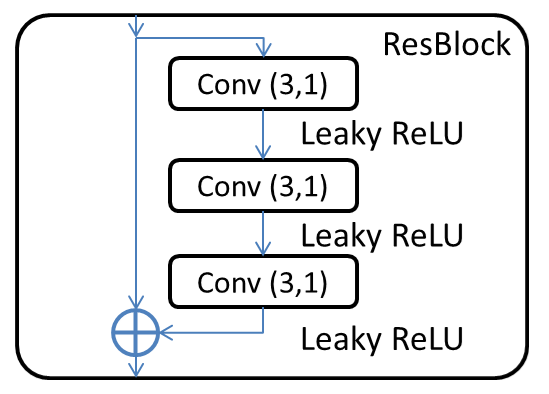}}
\caption{Structure of a residual block}
    \label{resblock}
\end{center}
\end{figure}

\autoref{resblock} depicts the structure of the used residual block. 
It consists of three stacked $3 \times 1$ convolutions.
A leaky rectified linear unit is the activation function, \cite{Maas13rectifiernonlinearities}.
The leaky ReLU also serves as a shortcut connection. 
Our observation is that batch normalization improves the convergence speed dramatically.
This aligns with similar results from \cite{pmlr-v37-ioffe15} and \cite{He2015},
However, at the same time, it hampers the network's ability to learn "deeper" patterns. 
Other works using deep learning for financial data do not use batch normalization.
Examples of this are \cite{zhang2019deeplob}, \cite{Tsantekidis2017}, and \cite{feng2019temporal}. 
We assume that batch normalization might be a "smoother."
As a consequence, it might affect deeper patterns in the financial time-series data. 

\begin{figure}[ht!]
%\vskip 0.2in
\begin{center}
\centerline{\includegraphics[width=0.9\columnwidth]{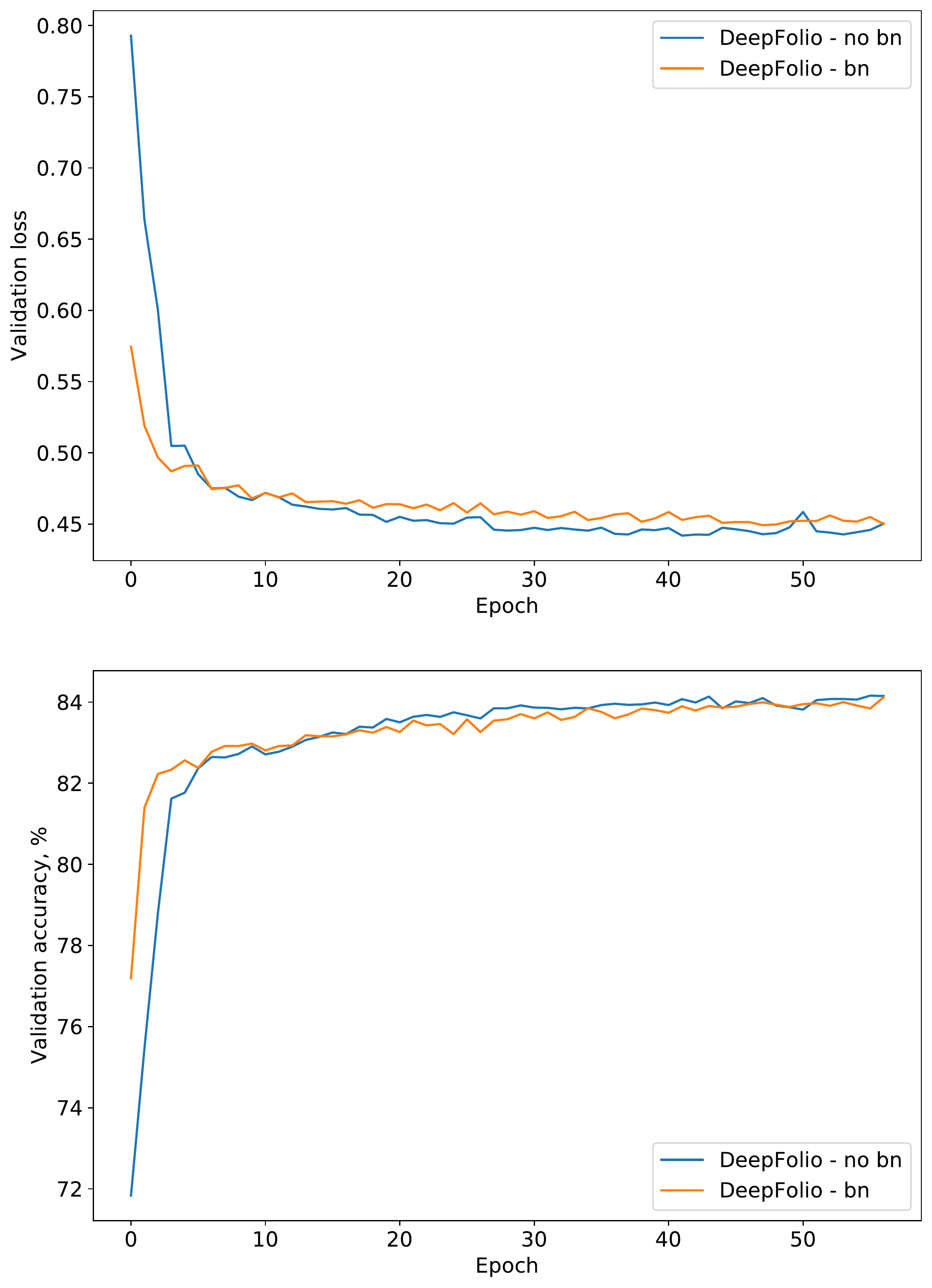}}
\caption{Comparison of validation loss (upper) and validation accuracy (lower) for DeepFolio with batch norms and without on FI-2010 dataset with prediction horizon $k = 1$}
    \label{batchnorm}
\end{center}
\end{figure}

\autoref{batchnorm} presents a comparison of two networks using the same dataset.
One has batch normalization, and the other does not. 
Negatively, the loss is higher in the validation dataset with batch normalization. 
Hence, we do not use it in the residual blocks.

\begin{figure}
%\vskip 0.2in
\begin{center}
\centerline{\includegraphics[width=0.9\columnwidth]{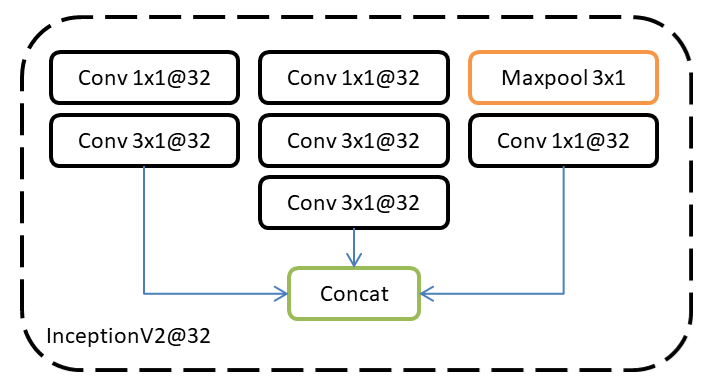}}
\caption{Structure of the "inception v2" module}
    \label{inceptionv2}
\end{center}
\end{figure}

In \autoref{inceptionv2}, we use the architecture "inception v2".
One can consider it as an alternative to the canonical inception block. 
\cite{szegedy2015rethinking} proposed it first.
The authors replace the $5 \times 5$ kernel with two consecutive smaller $3 \times 3$ kernels. 
This approach improves metrics and computational speeds.

\begin{figure}
%\vskip 0.2in
\begin{center}
\centerline{\includegraphics[width=0.5\columnwidth]{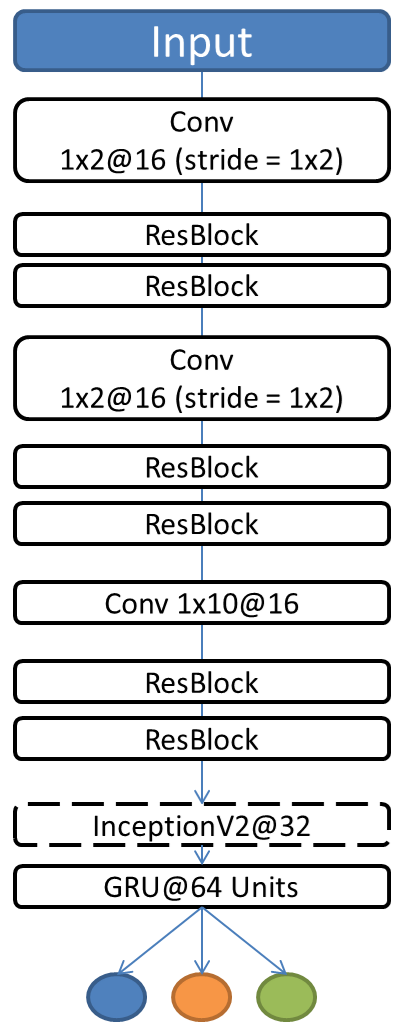}}
\caption{The network architecture of DeepFolio}
    \label{architecture}
\end{center}
\end{figure}

For most tasks, the gated recurrent units (GRU) performs on par with the LSTM.
We make this conclusion based on empirical observations.
Our conclusion arises from a numerical comparison of GRU versus LSTM.
However, GRUs offer additional benefits.
They have a more straightforward structure.
It allows them to generalize better in cases of limited data. 
Our architectural choices are visible in \autoref{architecture}.
We present the full architecture of the ResCNN+GRU module of DeepFolio.

\begin{figure}
%\vskip 0.2in
\begin{center}
\centerline{\includegraphics[width=\columnwidth]{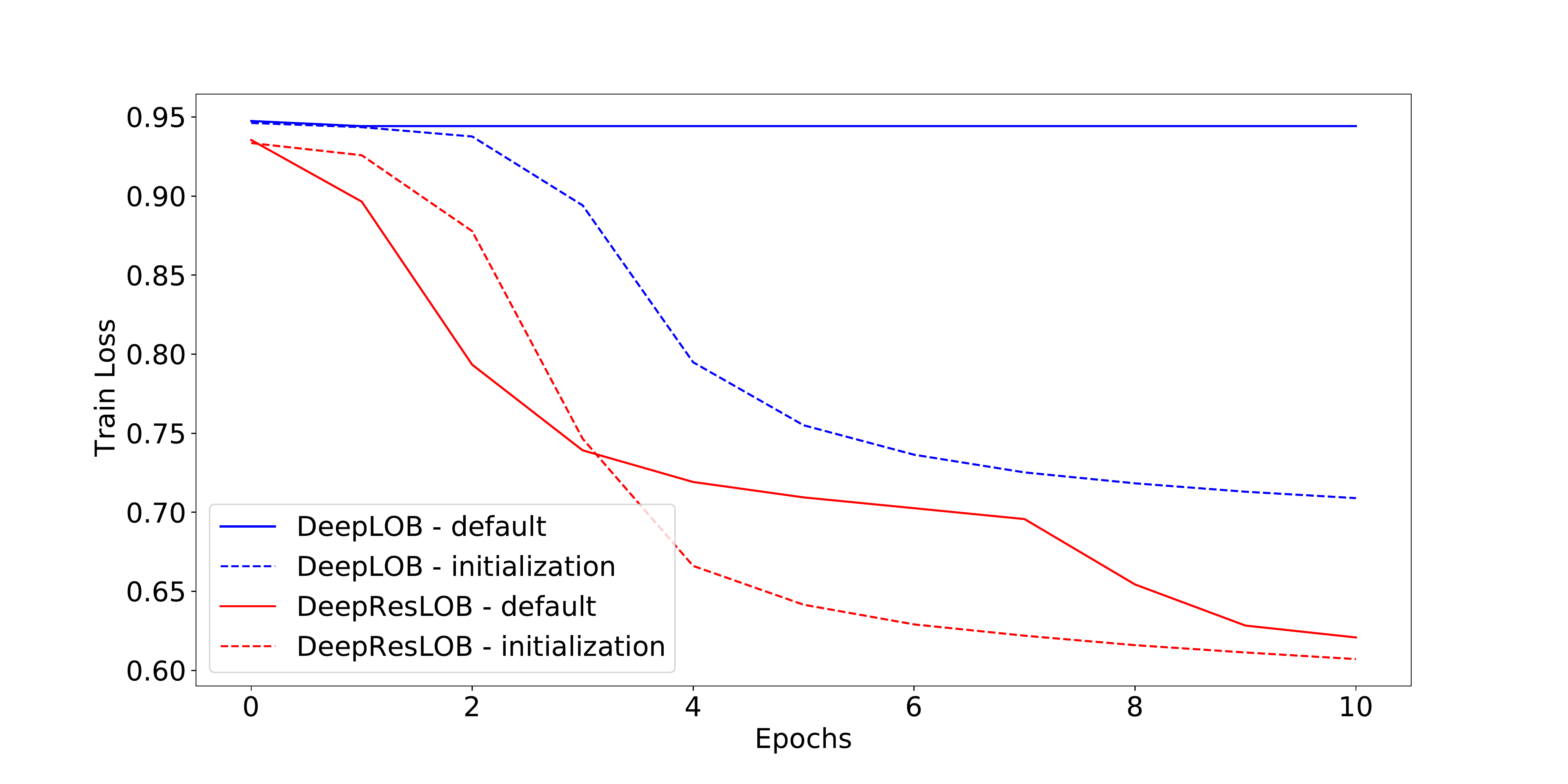}}
\caption{Training loss curves for both models with different initial weights allocations on the FI-2010 dataset with prediction horizon $k = 1$}
    \label{exp1}
\end{center}
\end{figure}

A problem of \cite{zhang2019deeplob} is its initial weight allocation dependency. 
In \autoref{exp1}, we can see that our ResCNN+GRU module solves it. 
It is mostly indifferent to the weight allocations.
Further, it trains well for both cases.

\begin{figure}
%\vskip 0.2in
\begin{center}
\centerline{\includegraphics[width=\columnwidth]{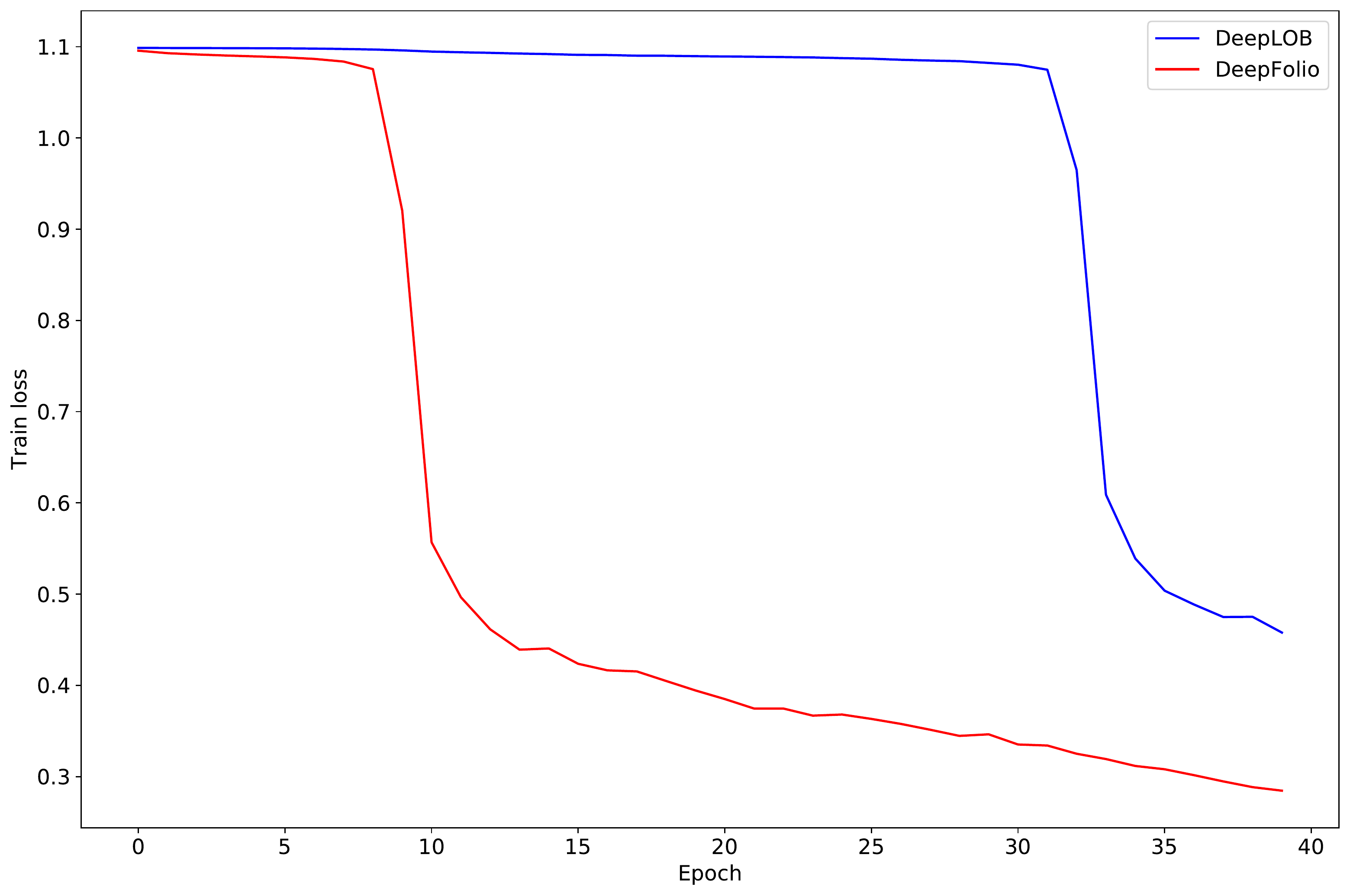}}
\caption{Comparison of training losses of DeepLOB and ResCNN+GRU of DeepFolio on the Bitcoin dataset with prediction horizon $k = 1$}
    \label{exp2}
\end{center}
\end{figure}

Another problem of \cite{zhang2019deeplob} is noticeable in the crypto-asset dataset. 
We run both models for the dataset of the crypto-asset BTC.
Our prediction horizon is $k = 1$ to see the performance of DeepFolio.
DeepLOB takes nearly 30+ epochs for the loss to start dropping.
On the other side, our model starts training at around epochs 8-9. 
Visually, we confirm in \autoref{exp2} that the problem disappears.

\subsection{Portfolio optimization model}
The predicted labels of DeepFolio are convenient for the development of trading strategies.
However, we go a step further of price prediction and trading strategies in this work. 
Our objective is to generate investment portfolios of crypto-assets.
For this, we build a crypto-asset portfolio consisting of 4 crypto-assets. 
We also perform weight rebalancing every 50 minutes.
Rather than strictly building portfolios with historical data, we use our predictions. 
It results in picking the period for rebalancing.
The reasoning is that it is less frequent than a predictive horizon of $k = 1$, i.e., 5 minutes.
Nevertheless, we still maintain reliable performance. 
To allocate portfolio weights, the model essentially has a two-step structure.
First, we feed the input data to the LSTM network.
Then, we pass the LSTM outputs through the fully connected layer with softmax activation. 
LSTMs are very efficient tools for modeling time series and especially financial data. 
Our innovation is that we use price movement labels to perform rebalancing.
Traditionally, the literature works with the price and returns history.

The following algorithm performs the training scheme.
Fist, the input for LSTM layers with 64 units consists of price movement indicators.
These are the labels from the ResCNN+GRU module. 
In our case, the period consists of 50 minutes. 
Second, we pass the predictions through softmax.
With this, we can get portfolio weights and use them to optimize the objective function.
Third, we run an Adam optimizer with a learning rate of 0.001.
We use this to train our network and set the batch size to 64. 
Fourth, after we train the network, we use the input predicted labels.
The ResCNN+GRU module generates them for intervals of 50 minutes.
As a result, we obtain the portfolio weights rebalanced. 
Fifth, we move ahead to the next 50-minutes interval. 
Again, we feed the input with predictions from the ResCNN+GRU module and update weights.
Finally, we repeat this process for the whole test set.
In this work, we evaluate two different loss functions:

\begin{enumerate} 
\item Maximization of Sharpe ratio, proposed in \cite{zhang2019deeplob}:
\begin{equation*}
L_\text{SR} = -\frac{\mathbb{E}[R]}{\text{std}(R)}
\end{equation*}
where:
\begin{equation*}
\mathbb{E} [R] = \mathbb{E} \left[ \sum^n_{i=1} w_{i,t-1} r_{i,t} \right]
\end{equation*}

and $r_{i,t} = (p_{i,t} - p_{i,t-1}) /  p_{i,t-1}$ is the return of the asset $i$.
$std$ is the standard deviation. 
Sharpe ratio is essentially a form of risk-adjusted returns.
It assesses the "optimality" of the portfolio. 
Portfolios with a higher Sharpe ratio are considered more optimal.

\item Minimization of portfolio volatility (risk):
\begin{equation*}
L_\text{V} = \text{std}(R)
\end{equation*}

It corresponds to the minimization of volatility.
This is equal to reducing portfolio risks.

\end{enumerate}

\section{Experiments}\label{sec:results}
We evaluate our model and compare its performance with the state-of-the-art.
Besides, we also consider two more baseline models. 
They are a CNN, \cite{Tsantekidis2017}, and an LSTM, \cite{Tsantekidis2017lstm}. 
For DeepLOB, we follow the indications in its respective publication strictly. 
To train the ResCNN+GRU module of DeepFolio, we use an Adam optimizer. 
We set its learning rate at 0.01, and $\epsilon$ to 1. 
To avoid overfitting, we apply early stopping with checkpointing. 
It saves the model weights each time. 
Our performance metrics are accuracy for FI-2010 and F1 score for the crypto-asset. 
On each iteration, we seek to improve them on the validation set.
If we do not observe changes after 20 epochs, the training stops. 
$L_2$-normalization helps us tackle overfitting.
It is especially relevant for the ResCNN+GRU module of DeepFolio.
Sometimes, it can overfit the training data. 
For example, validation loss starts to grow steadily.
We suppose that this is due to the deeper architecture with more parameters.

\subsection{FI-2010}
For the FI-2010 dataset, we divide ten days of this dataset into three parts. 
We use seven days for training and two days for validation.
The remaining days serve us as a training metric. 
We use 40 features from the dataset. 
They account for the ten levels of ask prices, bid prices, and quantities. 
The last five features are labels.
Respectively, they account for the prediction horizons $k = 1, 2, 3, 5$, and $10$. 
We use only $k = 1, 5, 10$ for comparison. 
These labels represent three different horizons.
They are short-term predictions, mid-term predictions, and long-term predictions. 
We also employ a sliding time window of length $T = 100$ with a batch size equal to 64.
The input to the network has a size (64, 1, 100, 40).
In this case, the second dimension is an auxiliary "channel" dimension.

In \autoref{Tab: FI-2010}, we see the benefits of our model. 
Both DeepLOB and DeepFolio massively outperform both baseline models.
The difference grows further as the length of the prediction horizon, $k$, grows.
DeepFolio also outperforms DeepLOB on all metrics. 
The performance gap between these two models also grows with the length of $k$. 
The architecture of DeepFolio captures the long-term relations in the data better.

\begin{table}
\centering
\caption{\textsc{Experimental results for FI-2010 benchmark dataset with different prediction horizons $k$}\label{Tab: FI-2010}}
\resizebox{0.48\textwidth}{!}{%
\begin{tabular}{lcccc}
\hline
Model & Accuracy \%      & Precision \%     & Recall \%        & F1 \%            \\ \hline
\multicolumn{5}{c}{Prediction horizon k   = 10}                                 \\ \hline
CNN            & 41,23\%          & 44,54\%          & 45,89\%          & 38,40\%          \\ 
LSTM           & 38,31\%          & 25,12\%          & 33,32\%          & 18,63\%          \\ 
DeepLOB        & 77.39\%          & 80.72\%          & 77.39\%          & 77.11\%          \\ 
DeepFolio          & \textbf{79.51\%} & \textbf{82.18\%} & \textbf{79.51\%} & \textbf{79.22\%} \\ 
\hline
\multicolumn{5}{c}{Prediction horizon k   = 5}                                  \\ \hline
CNN            & 58,11\%          & 50,76\%          & 55,25\%          & 50,67\%          \\ 
LSTM           & 50,60\%          & 16,87\%          & 33,33\%          & 22,40\%          \\ 
DeepLOB        & 74.26\%          & 77.58\%          & 74.26\%          & 73.7\%           \\ 
DeepFolio          & \textbf{75.03\%} & \textbf{77.66\%} & \textbf{75.03\%} & \textbf{74.51\%} \\ 
\hline
\multicolumn{5}{c}{Prediction horizon k = 1}                                   \\ \hline
CNN            & 77,88\%          & 75,53\%          & 60,56\%          & 65,12\%          \\ 
LSTM           & 66,93\%          & 22,31\%          & 33,33\%          & 26,73\%          \\ 
DeepLOB        & 81.8\%           & 83.02\%          & 81.8\%           & 80.88\%          \\
DeepFolio          & \textbf{82.44\%} & \textbf{83.98\%} & \textbf{82.44\%} & \textbf{81.29\%} \\ \hline
\end{tabular}
}
\end{table}

\subsection{Crypto-asset dataset}
We consider two different cases for the crypto-asset dataset. 
The first setup is a conventional one. 
We train a separate network for each crypto-asset.
Then, we validate and test only on the respective crypto-assets.
The second approach combines three crypto-assets into one dataset.
They are BTC, LTC, and ETH.
We do the training on this combined dataset. 
Separately, we perform testing on each crypto-asset.
That way, we can assess the models' ability to generalize. 
Also, we intentionally hold out Ripple (XRP) entirely.
We aim to additionally back-test the models.
We want to evaluate their generalization ability to do transfer learning.
For both approaches, we use a sliding time window of $T = 60$ and a batch size of 64. 
For the first case, we employ a 70-15-15 split of the datasets.
Respectively, we use 70\% for training and 15\% for validation and test.
An additional characteristic is that the datasets are unbalanced. 
Hence, we focus on the weighted F1 score to assess the performance of the models.

\begin{figure*}
\begin{center}
\centerline{\includegraphics[width=0.9\textwidth]{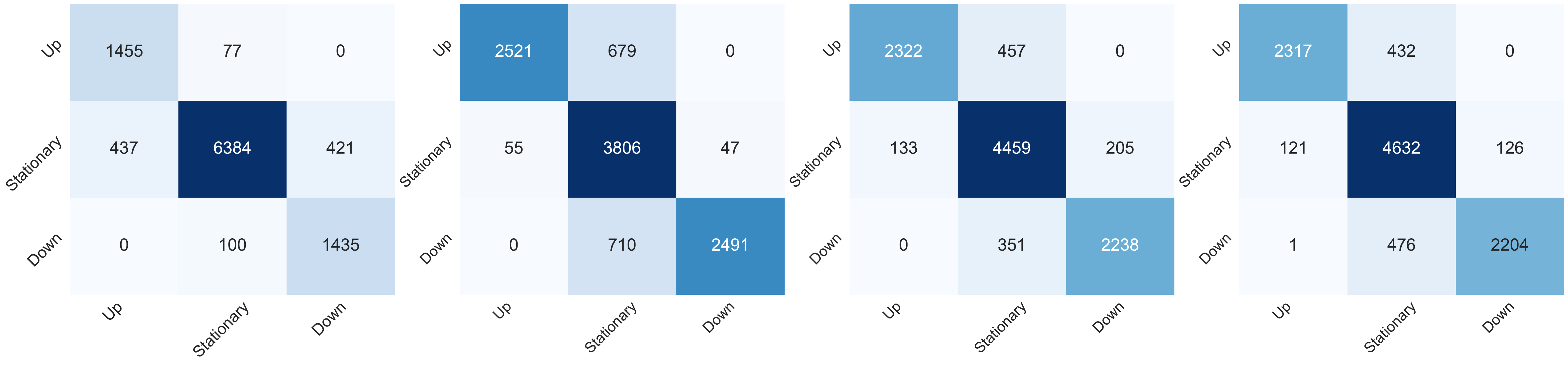}}
\caption{Confusion Matrix for crypto-assets. Prediction horizon ($k$) equals 1. From left to right: "BTC", "LTC", "ETH", "XRP"}
    \label{Fig: confusion k1}
\end{center}
\end{figure*}

\begin{figure*}
\begin{center}
\centerline{\includegraphics[width=0.9\textwidth]{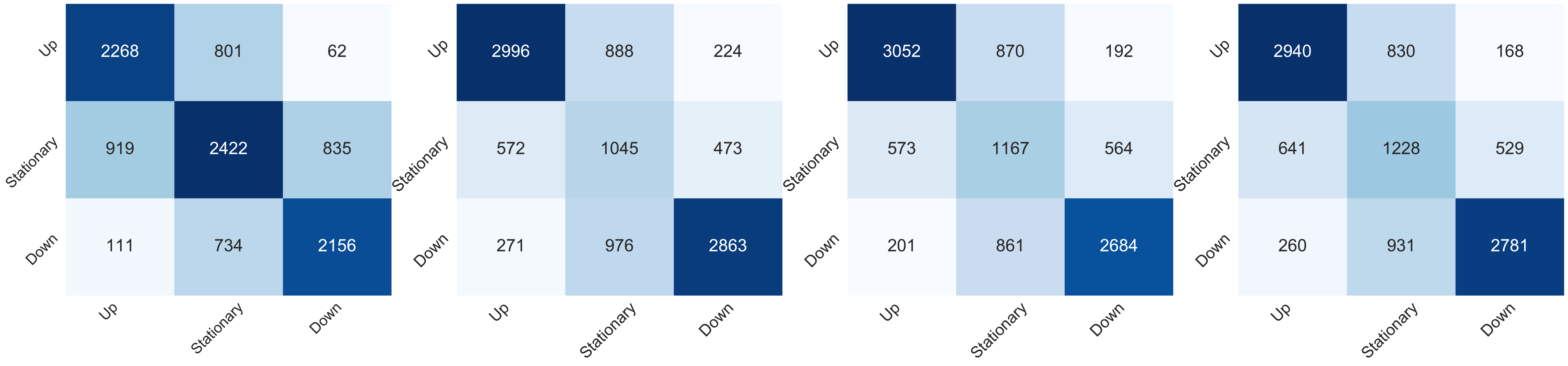}}
\caption{Confusion Matrix for crypto-assets. Prediction horizon ($k$) equals 5. From left to right: "BTC", "LTC", "ETH", "XRP"}
    \label{Fig: confusion k5}
\end{center}
\end{figure*}

\begin{figure*}
\begin{center}
\centerline{\includegraphics[width=0.9\textwidth]{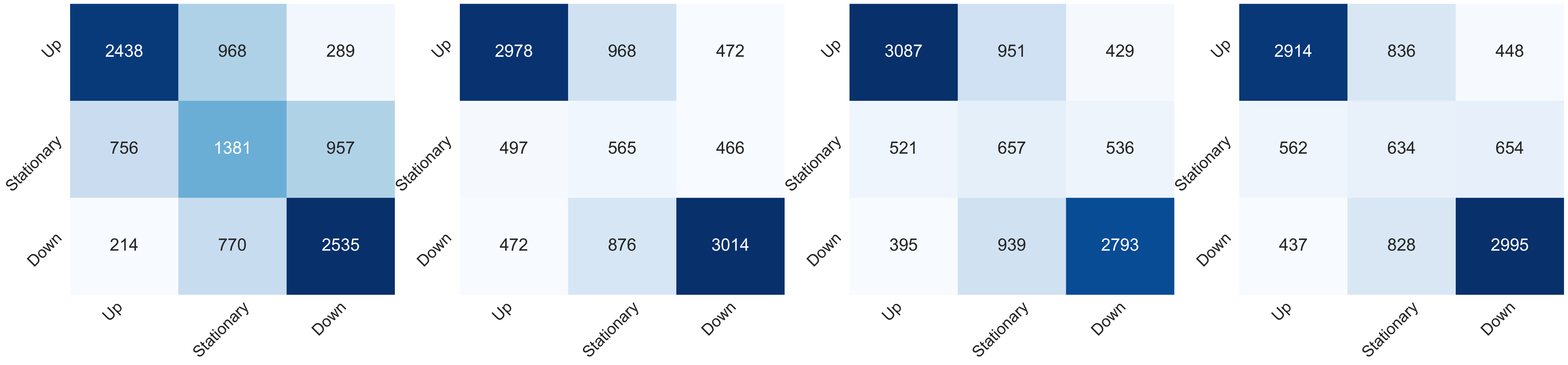}}
\caption{Confusion Matrix for crypto-assets. Prediction horizon ($k$) equals 10. From left to right: "BTC", "LTC", "ETH", "XRP"}
    \label{Fig: confusion k10}
\end{center}
\end{figure*}

\begin{figure*}
\begin{center}
\centerline{\includegraphics[width=0.9\linewidth]{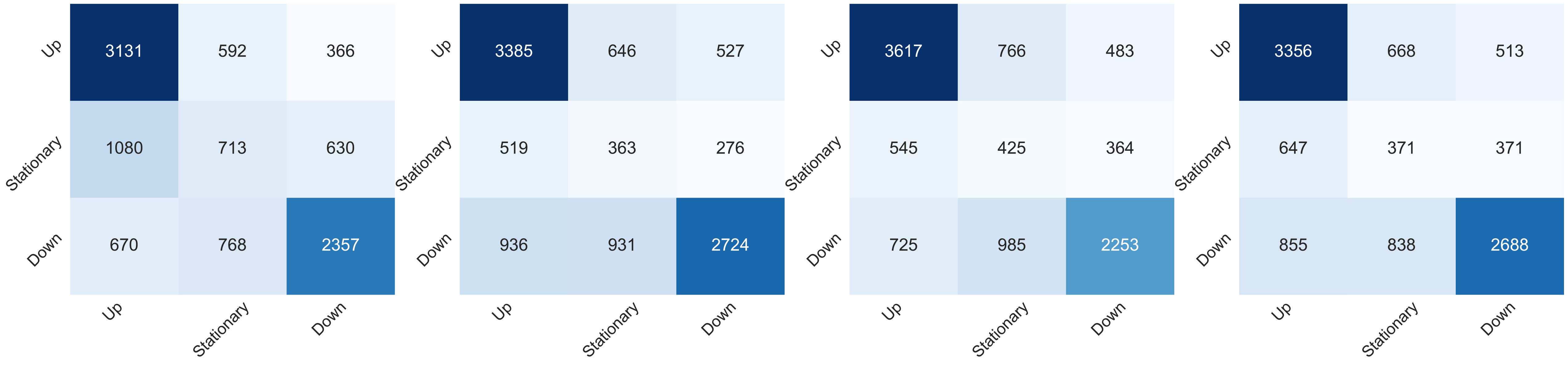}}
\caption{Confusion Matrix for crypto-assets. Prediction horizon ($k$) equals 20. From left to right: "BTC", "LTC", "ETH", "XRP"}
    \label{Fig: confusion k20}
\end{center}
\end{figure*}

In \autoref{Tab: Exp crypto}, we appreciate that both DeepLOB and DeepFolio outperform.
The baselines show worse performance by a large margin on all metrics. 
When we move to longer prediction horizons, it becomes especially evident.
Rapidly, the metrics of baseline methods start dropping. 
DeepLOB and DeepFolio also experience a decrease in metrics.
Nevertheless, it is not as severe as the baseline models. 
While directly comparing DeepFolio and DeepLOB, we can see that DeepFolio outperforms. 
It gets superior scores across all metrics. 
However, the gap between them is narrow. 
To better investigate the results, we provide the confusion matrices.
They are available for the four prediction horizons in \autoref{Fig: confusion k1} for $k$ = 1.
In \autoref{Fig: confusion k5}, they have $k$ = 5.
We also present two additional matrices for further horizons.
In \autoref{Fig: confusion k10}, it is $k$ = 10.
$k$ = 20 is in \autoref{Fig: confusion k20}.

\begin{table}
\centering
\caption{\textsc{The results of experiments on cryptoassets with the first setup for different prediction horizons $k$}\label{Tab: Exp crypto}}
\resizebox{0.48\textwidth}{!}{%
\begin{tabular}{lllll}
\hline
Model & Accuracy \%      & Precision \%     & Recall \%        & F1 \%                 \\ \hline
\multicolumn{5}{c}{Prediction horizon k   = 1}                                                \\ \hline
CNN             & 77,88\%          & 75,53\%          & 60,56\%          & 65,12\%               \\
LSTM            & 68,68\%          & 22,89\%          & 33,33\%          & 27,14\%               \\ 
DeepLOB         & 81.02\%          & \textbf{89.21\%} & 81.02\%          & 81.89\%               \\ 
DeepFolio           & \textbf{84.84\%} & 89.11\%          & \textbf{84.84\%} & \textbf{84.32\%}      \\ \hline
\multicolumn{5}{c}{Prediction horizon k   = 5}                                                \\ \hline
CNN             & 76,32\%          & 38,45\%          & 42,78\%          & 40,45\%               \\ 
LSTM            & 40,02\%          & 13,34\%          & 33,33\%          & 19,05\%               \\
DeepLOB         & 64.68\%          & 69.19\%          & 64.68\%          & 65.04\%               \\ 
DeepFolio           & \textbf{65.17\%} & \textbf{69.73\%} & \textbf{65.17\%} & \textbf{65.48\%}      \\ \hline
\multicolumn{5}{c}{Prediction horizon k   = 10}                                               \\ \hline
CNN             & 23,37\%          & 26,35\%          & 17,88\%          & 20,41\%               \\ 
LSTM            & 13,51\%          & 17,15\%          & 9,77\%           & 11,37\%               \\ 
DeepLOB         & 59.81\%          & 63.15\%          & 59.81\%          & 58.32\%               \\ 
DeepFolio           & \textbf{60.28\%} & \textbf{66.6\%}  & \textbf{60.28\%} & \textbf{60.87\%}      \\ \hline
\multicolumn{5}{c}{Prediction horizon k   = 20}                                               \\ \hline
CNN             & 21,60\%          & 20,27\%          & 14,98\%          & 15,91\%               \\ 
LSTM            & 14,40\%          & 9,95\%           & 9,93\%           & 9,94\%                \\
DeepLOB         & 53.09\%          & 67.33\%          & 53.09\%          & 55.63\%               \\ 
DeepFolio           & \textbf{55.43\%} & \textbf{67.43\%} & \textbf{55.43\%} & \textbf{57.91\%}      \\ \hline
\end{tabular}
}
\end{table}

For the second setup, we split the dataset in the following way. 
First, we take each crypto-asset from the (BTC, LTC, ETH) trio separately.
Then, we perform an 80-10-10 train-validation-test split. 
After that, we concatenate the train parts of the crypto-assets.
With this, we form a single dataset.
We repeat the same process for the validation, while we keep the test sets separate. 
The main goal of this setup is to check whether models can extract general LOB patterns.
Our inspiration is the work of \cite{sirignano2018universal}. 
To further test the networks' ability, we perform transfer learning.
We select the XRP crypto-asset for this task.
We feed it to the entire dataset into models that did not previously see the XRP data.
For this setup, we exclude baseline models. 
Their performance is limited, even when dealing with individual crypto-assets. 
Thus, we focus on DeepLOB and DeepFolio, primarily.

We look at \autoref{Tab: transfer learning}.
It seems that both models have strong generalizing abilities.
However, DeepFolio outperforms in the majority of the cases.
The gaps this time are higher at about 2-3 \% on average. 
Transfer learning results are also robust.
It means that neural networks are indeed capable of learning the general LOB patterns.
They do not merely adapt to the data. 
Overall, in both setups, we can see that DeepFolio outperforms.

\begin{table}
\centering
\caption{\textsc{Results for the transfer learning setup for different prediction horizons $k$ using the crypto-asset dataset }\label{Tab: transfer learning}}
\resizebox{0.48\textwidth}{!}{%
\begin{tabular}{lllll}
\hline
Model                                  & Accuracy \%           & Precision \%          & Recall \%             & F1 \%                 \\ \hline
\multicolumn{5}{c}{Prediction horizon k   = 1}                                                                                                 \\ \hline
DeepLOB                                          & 86.69\%               & 91.67\%               & 86.69\%               & 87.44\%               \\
DeepFolio                                            & \textbf{89.9\%}       & \textbf{93.14\%}      & \textbf{89.9\%}       & \textbf{90.4\%}       \\ \hline
\multicolumn{5}{c}{Prediction horizon k   = 5}                                                                                                 \\ \hline
DeepLOB                                          & 66.43\%               & \textbf{70.04\%}      & 66.43\%               & \textbf{66.97\%}      \\ 
DeepFolio                                            & \textbf{66.46\%}      & 69.1\%                & \textbf{66.46\%}      & 66.75\%               \\ \hline
\multicolumn{5}{c}{Prediction horizon k   = 10}                                                                                                \\ \hline
DeepLOB                                          & 56.96\%               & \textbf{71.53\%}      & 56.96\%               & 58.39\%               \\ 
DeepFolio                                            & \textbf{61.57\%}      & 66.35\%               & \textbf{61.57\%}      & \textbf{62.33\%}      \\ \hline
\multicolumn{5}{c}{Prediction horizon k   = 20}                                                                                                \\ \hline
DeepLOB                                          & 57.6\%                & 69.29\%               & 57.6\%                & 57.6\%                \\
DeepFolio                                            & \textbf{59.81\%}      & \textbf{67.08\%}      & \textbf{59.81\%}      & \textbf{60.75\%}      \\ \hline
\end{tabular}
}
\end{table}

\subsection{Portfolio}
To evaluate our portfolio model performance, we estimate the portfolio value using \cite{Jiang2017} and define it as
\begin{equation*}
    p_t = p_{t-1} \frac{r_t}{r_{t-1}} w_{t-1}
\end{equation*}

where $p_{t-1}$ is the portfolio value at the beginning of period $t$.
$r_t$ corresponds to prices vector at time $t$.
Meanwhile, $w_{t-1}$ is the portfolio weight vector at the beginning of period $t$. 
We rebalance every 50 minutes and do not consider transaction costs.

\begin{figure}
\begin{center}
\centerline{\includegraphics[width=\columnwidth]{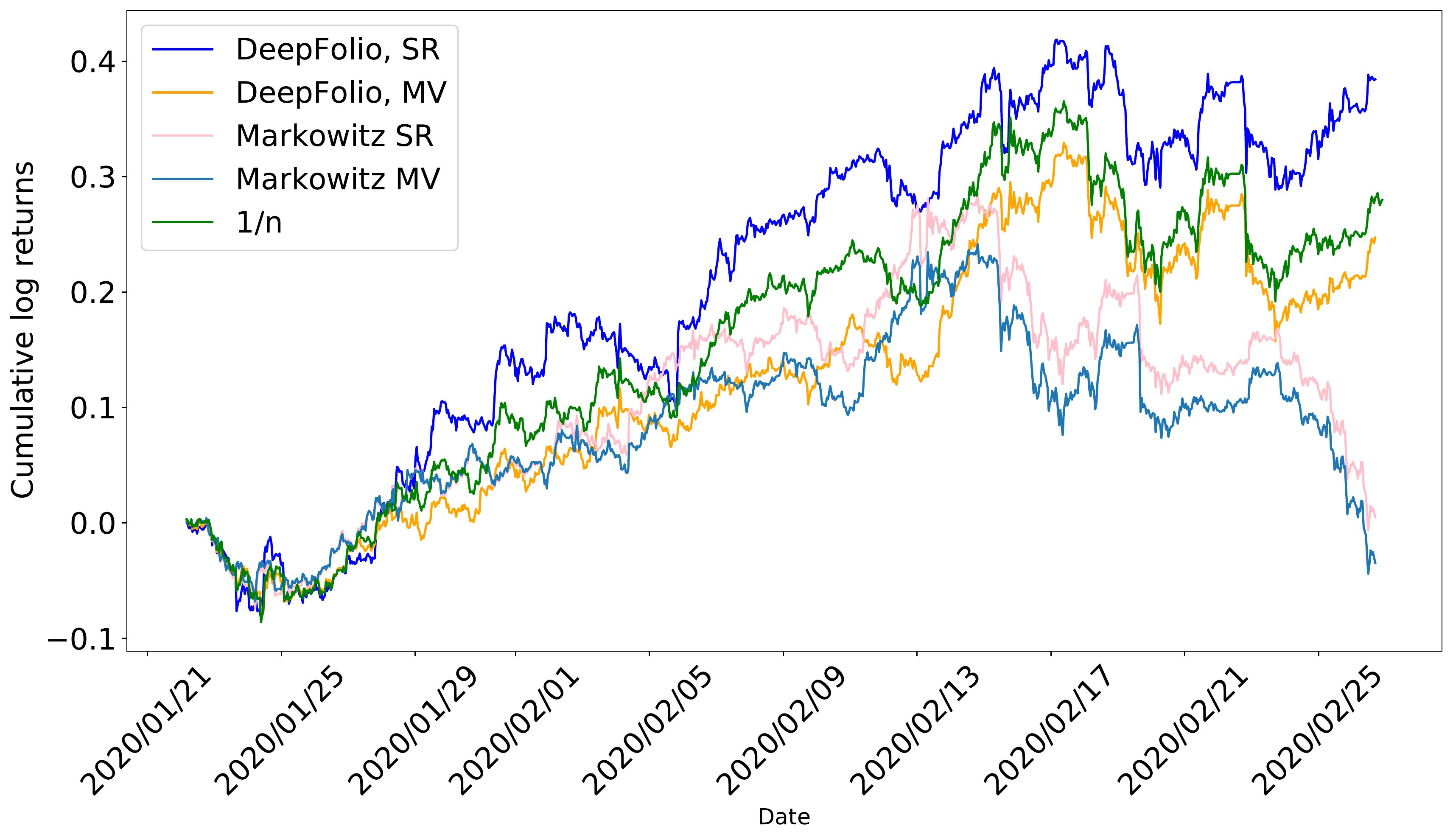}}
\caption{Cumulative returns in logarithmic scale for the various portfolio strategies}
    \label{Fig: portfolio values}
\end{center}
\end{figure}

In \autoref{Fig: portfolio values}, we see the various portfolio strategies. 
It displays the cumulative log-returns.
Here, $1/n$ is the equal-weights naive portfolio. 
Markowitz SR corresponds to the Markowitz model with Sharpe Ration.
Similarly, Markowitz MV has a mean-variance. 
DeepFolio SR uses the Sharpe Ratio as a loss function.
However, for DeepFolio MV, we have volatility, instead. 
DeepFolio with Sharpe Ratio has the best performance on the test dataset. 
Moreover, the testing period starts around February 2020.
In this period, the crisis induced by COVID-19 hits the global markets.

\autoref{Tab: Exp benchmark} presents a global comparison of results. 
We want a full understanding of each method's performance.
For this, we compare the following parameters.
First, we consider the expected and mean returns.
Second, we look at the standard deviation of portfolio returns, the Sharpe Ratio.
Third, we have the ratio between positive and negative returns for the test period. 
We can see that all reallocation strategies work well. 
Nevertheless, DeepFolio with Sharpe Ratio shows the best values for all parameters.
The only exception is in the case of the standard deviation.

\begin{table*}
\centering
\caption{\textsc{Experimental results for different algorithms on crypto-assets dataset}\label{Tab: Exp benchmark}}
\vskip 0.1in
\resizebox{\textwidth}{!}{%
\begin{tabular}{crrrrr}
\hline
\multicolumn{1}{l}{}   & Expected Return & Mean Return & Standard Deviation      & Sharpe Ratio & +/-      \\
\hline
Markowitz, SR	       & 1.152323        & 0.016102    & 0.006193          & 0.025998             & 1.108280
 \\         
Markowitz, MV           & 1.159850                 & 0.016629             & \textbf{0.005987}          & 0.027774    & 1.081761              \\
1/n & 1.281730                 &0.027101            & 0.006736          & 0.040234	              & \textbf{1.130901} 
\\
\hline
DeepFolio, SR      & \textbf{1.467931}                 &   \textbf{0.040126}            & 0.007561          & \textbf{0.053069	}               & 1.118012    \\
DeepFolio, MV      & 1.280971                 & 0.025986             & 0.006225          & 0.041746	             & 1.113636           \\

\hline
\end{tabular}
}
\end{table*}

\section{Conclusions and discussion}\label{sec:conclussion}
We propose DeepFolio to address problems in the state-of-the-art.
Our model surpasses its performance on the benchmark dataset.
We observe similar behavior for the crypto-asset dataset.
It is despite the latter being more scarce and favoring smaller models.
We also show that DeepFolio is capable of learning general patters in the LOB data.
It does not merely adapt to the data at hand.
We demonstrate it through transfer learning on a previously unseen crypto-asset.
We generate price movement predictions from LOBs.
With them, we prove that they as well can be used for short-term portfolio allocation.
We bestow these portfolios with rebalancing strategies.
Such an approach overcomes the pitfalls of classical methods of portfolio optimization.
Also, we test the model with two different loss functions.
They are the maximization of Sharpe ratio and volatility. 
Extensive tests show that DeepFolio with Sharpe Ratio performs the best.
It outperforms all other approaches.
Portfolio managers can use the results of this work for a myriad of assets.
For assets with high liquidity, we expect a better performance.
They are less prone to stochastic fluctuations.
In conclusion, our approach serves as a building block for an automated portfolio building and optimization framework.

\bibliographystyle{IEEEtran}
\bibliography{bib}

\clearpage

\end{document}